\documentclass[twoside,11pt]{article}
\usepackage[specialissue, accepted]{melba}

\usepackage{mwe} 

\usepackage{amsmath,amsfonts}
\usepackage{booktabs} 
\usepackage[utf8]{inputenc}
\usepackage{multirow}
\usepackage{bigstrut}
\usepackage{amssymb}

\melbaid{2024:017}  
\doi{https://doi.org/10.59275/j.melba.2024-489g}
\melbaauthors{Wang et al.}  
\volume{2}
\firstpageno{1030}  
\melbayear{2024}  
\datesubmitted{04/2024}  
\datepublished{08/2024}  

\melbaspecialissue{MICCAI 2023 LNQ challenge special issue}
\melbaspecialissueeditors{Steve Pieper, Erik Ziegler, Tawa Idris, Bhanusupriya Somarouthu, Reuben Dorent, Gordon Harris, Ron Kikinis}

\ShortHeadings{Label-efficient Learning for Partial Instance Annotations}{Wang et al.}

\title{Weakly Supervised Lymph Nodes Segmentation Based on Partial Instance Annotations with Pre-trained Dual-branch Network and Pseudo Label Learning}

\author{%
	\firstname Litingyu \surname Wang \email litingyuwang@gmail.com \\
    \addr University of Electronic Science and Technology of China, Chengdu, China
    \AND
    \firstname Yijie \surname Qu \\
    \addr University of Electronic Science and Technology of China, Chengdu, China
    \AND
    \firstname Xiangde \surname Luo \\
    \addr University of Electronic Science and Technology of China, Chengdu, China \\
    \addr Shanghai AI Laboratory, Shanghai, China
    \AND
    \firstname Wenjun \surname Liao \\
    \addr University of Electronic Science and Technology of China, Chengdu, China \\
    \addr Department of Radiation Oncology, Sichuan Cancer Hospital \& Institute, Sichuan Cancer Center, Chengdu, China
    \AND
    \firstname Shichuan \surname Zhang \\
    \addr University of Electronic Science and Technology of China, Chengdu, China \\
    \addr Department of Radiation Oncology, Sichuan Cancer Hospital \& Institute, Sichuan Cancer Center, Chengdu, China
    \AND
    \firstname Guotai \surname Wang\orcid{0000-0002-8632-158X} \email guotai.wang@uestc.edu.cn \\
    \addr University of Electronic Science and Technology of China, Chengdu, China \\
    \addr Shanghai AI Laboratory, Shanghai, China
}

\begin{document}
\maketitle

\begin{abstract}
	Assessing the presence of potentially malignant lymph nodes aids in estimating cancer progression, and identifying surrounding benign lymph nodes can assist in determining potential metastatic pathways for cancer. For quantitative analysis, automatic segmentation of lymph nodes is crucial. However, due to the labor-intensive and time-consuming manual annotation process required for a large number of lymph nodes, it is more practical to annotate only a subset of the lymph node instances to reduce annotation costs. In this study, we propose a pre-trained Dual-Branch network with Dynamically Mixed Pseudo label (DBDMP) to learn from partial instance annotations for lymph nodes segmentation. To obtain reliable pseudo labels for lymph nodes that are not annotated, we employ a dual-decoder network to generate different outputs that are then dynamically mixed. We integrate the original weak partial annotations with the mixed pseudo labels to supervise the network. To further leverage the extensive amount of unannotated voxels, we apply a self-supervised pre-training strategy to enhance the model's feature extraction capability. Experiments on the mediastinal Lymph Node Quantification (LNQ) dataset demonstrate that our method, compared to directly learning from partial instance annotations, significantly improves the Dice Similarity Coefficient (DSC) from 11.04\% to 54.10\% and reduces the Average Symmetric Surface Distance (ASSD) from 20.83~$mm$ to 8.72~$mm$.
	The code is available at~\url{https://github.com/WltyBY/LNQ2023_training_code.git}
\end{abstract}



\section{Introduction}
	Lymph node segmentation is essential in various medical applications, particularly in the diagnosis, staging and treatment planning of diseases such as cancer~\citep{bouget2023mediastinal}. By segmenting lymph nodes and monitoring factors, such as their size and shape, clinicians can track disease progression and formulate treatment plans~\citep{li2020deep}. Deep learning methods have shown promising results in medical image segmentation~\citep{chen2021transunet, isensee2021nnu}. However, obtaining the necessary pixel-level annotations is extremely time-consuming and labor-intensive~\citep{pathak2015constrained}. Therefore, leveraging weak annotations to train a deep learning model is highly desirable for reducing annotation costs.

    Weakly supervised learning can be extremely beneficial for training deep learning models in lymph node segmentation. Lymph nodes are distributed throughout the human body, often small in size, making it impractical to label all lymph nodes in a given area. In traditional weakly supervised tasks, labels for training segmentation models typically include image-level~\citep{fu2023cam}, box-level~\citep{oh2021background}, point-level~\citep{zhai2023pa} and scribble-level~\citep{luo2022scribble} annotations. Image-level annotations merely indicate the presence of the target in the image and lack detailed information on its shape, intensity, and location, potentially leading to subpar performance. Due to the subtle contrast between lymph nodes and surrounding tissues in Computed Tomography (CT) scans, providing supervision information around the boundary poses a challenge for box-level and point/scribble-level annotations. In contrast, this study adopts a different weak annotation strategy called partial instance annotation, where only a small subset of lymph nodes are annotated in a volume. This annotation method provides the model with more information about the target's size, shape and boundary than the other weak annotations. Additionally, compared to fully supervised segmentation datasets, the annotation cost is significantly reduced.
    
    Some researchers~\citep{bouget2023mediastinal, feulner2013lymph} have proposed annotating only a subset of lymph nodes for training, for example, only annotating lymph nodes with a high probability of disease based on size (minimum diameter value larger than 10~$mm$). However, some works still use fully supervised training procedures on datasets labeled in this manner. \cite{oda2018dense} utilized Fully Convolutional Networks (FCNs) for mediastinal lymph node detection and segmentation, which is trained on annotated lymph nodes and other anatomical structures to address data imbalance. \cite{bouget2023mediastinal} introduced anatomical prior knowledge during training to assist the model in distinguishing lymph nodes from similar surrounding structures.
    However, solely segmenting diseased lymph nodes is inadequate for clinical use, as both diseased and normal lymph nodes are essential for diagnosing and treating diseases. Diseased lymph nodes offer insights into affected areas, while nearby normal lymph nodes can indicate potential metastasis paths for cancer. Therefore, it is imperative to segment both diseased and normal lymph nodes, even if only a subset of instances are annotated for training purpose. 

    In this work, we propose a novel framework named pre-trained \textbf{D}ual-\textbf{B}ranch network with \textbf{D}ynamically \textbf{M}ixed \textbf{P}seudo labels (DBDMP) which integrates self- and weakly supervised learning concepts along with noisy label learning to train a segmentation model with partial instance annotations. To better improve the feature extraction capability of the model, we employ a self-supervised pre-training method called Model Genesis~\citep{zhou2021models}, which involves an image reconstruction task. Additionally, within the framework of weakly supervised learning, we utilize a noise-robust loss to enhance learning from partial instance annotations. Furthermore, to effectively leverage unlabeled pixels during training, we introduce a real-time pseudo label learning strategy. We dynamically mix the outputs from two decoders to obtain soft pseudo labels, which are more robust to noise compared to hard pseudo labels ~\citep{muller2019does}. Subsequently, we merge the original partial annotations with the mixed predictions, leveraging the complementary information between the two kinds of labels for robust learning. The main contributions of this work are summarized as follows:
    \begin{itemize}
        \item[$\bullet$] We propose a novel pseudo label generation strategy for learning from partial instance annotations for lymph node segmentation. By assigning pseudo labels to unannotated lymph nodes instead of directly treating them as background, our approach effectively enhances the segmentation model's recall and reduces false negatives. Furthermore, the utilization of soft pseudo labels is more noise-tolerant than hard pseudo labels, which makes the training process more robust.
        \item[$\bullet$] During the pseudo label learning stage, a consensus-aware Cross-Entropy loss is proposed. The weight of each pixel is determined by the consistency between the two predictions derived from the weakly supervised learning framework. This approach facilitates the gradual learning of newly predicted foreground voxels by the model while mitigating the risk of being misled by incorrect ones.
        \item[$\bullet$] We adopt Model Genesis to initialize model parameters, enhancing the model's capability to extract superior features and edge information through the reconstruction of corrupted images.
    \end{itemize}
    
    Our method was validated on the Mediastinal Lymph Node Quantification (LNQ) dataset, and promising results have been achieved. In the LNQ challenge held on MICCAI 2023, we secured the $4^{th}$ position without utilizing any additional datasets for training, while other participants used extra training sets. Furthermore, our final methods attained a Dice Similarity Coefficient (DSC) score of 54.10\% on the validation set and 55.44\% on the test set, which correspond to an improvement of 43.06 and 36.40 percentage points compared with supervised learning from the annotated instances only, respectively. It is worth noting that our best DSC on the test set is 57.36\%.

\vspace{-0.5em}
\section{Related Works}
\vspace{-0.5em}
	\paragraph{Lymph Nodes Segmentation}
	Numerous efforts based on traditional vision-based methods have been dedicated to lymph node detection and segmentation, including Marginal Space Learning (MSL)~\citep{barbu2011automatic} and atlas-based segmentation~\citep{stapleford2010evaluation}, etc. However, traditional methods may face challenges such as suboptimal performance or excessive computation time~\citep{zhao2020deep}. In recent years, deep learning has been applied to lymph node segmentation due to its outstanding performance in tasks such as image classification and segmentation. \cite{nogues2016automatic} presented a method for automatic segmentation of lymph node clusters in CT images using holistically-nested neural networks and structured optimization. \cite{bouget2019semantic} proposed a 2D pipeline that integrates the outputs of U-Net~\citep{ronneberger2015u} and Mask R-CNN~\citep{he2017mask} for segmentation and improves the performance with the instance detection. \cite{xu2021disegnet} introduced a Cosine-Sine loss function and a multi-scale Atrous Spatial Pyramid Pooling (ASPP) module to the SegNet~\citep{badrinarayanan2017segnet} architecture to address the voxel class imbalance and enhance multi-scale information. Although these methods have achieved success in lymph node segmentation, they all relied on fully annotated training datasets with high annotation costs.

    \paragraph{Label-efficient Learning}
	The objective of label-efficient learning is to reduce the cost and time required by the labeling process while achieving performance comparable to fully supervised methods, especially for image segmentation tasks where it is expensive and time-consuming to obtain dense annotations~\citep{shen2023survey}. Label-efficient learning techniques encompass semi-supervised learning~\citep{luo2022semi}, active learning~\citep{settles2009active}, weakly supervised learning~\citep{luo2022scribble} and noisy label learning~\citep{wang2020annotation}, among others. For example, \cite{lin2016scribblesup} utilized a graphical model that jointly propagates information from scribbles to unlabeled pixels based on superpixels~\citep{ren2003learning}. \cite{luo2022scribble} employed auxiliary branch to generate pseudo labels in real-time and used a specific loss function to expand the scribbled regions. To ensure robustness against inaccurate annotations in segmentation tasks, \cite{liu2022adaptive} enforced multi-scale cross-view consistency, and \cite{wang2020noise} introduced a noise-robust Dice loss. Compared to inaccurate annotations in existing noise-robust methods, partial instance learning has a larger degree of errors due to that most instances have been erroneously taken as the background. Furthermore, as the annotation type is different from the above weak annotations, existing weakly supervised segmentation methods cannot be directly used for learning from partial instance annotations. 
    
    \paragraph{Self-supervised Learning}
    Self-supervised learning serves as a mechanism for models to learn rich feature representations from unlabeled data, thereby reducing reliance on large labeled datasets. This is commonly achieved by designing a pretext task. \cite{gidaris2018unsupervised} designed a classification-based pretext task to predict discretized rotation angles of an input image. \cite{nogues2016automatic} implemented the self-supervised task by solving jigsaw puzzles. \cite{zhou2021models} introduced Model Genesis which reconstructs a corrupted input to its original state. Designs like Model Genesis allow the model to extract universal image features effectively. \cite{lei2021contrastive} proposed a novel contrastive learning approach, estimating the relative 3D offset between any pair of patches within the same volume. This method can perform well with just one-shot fine-tuning, while most other methods require fully supervised fine-tuning in the downstream task. However, in existing works, models trained by self-supervised learning are mainly fine-tuned with a small set of fully annotated images in downstream tasks, while applying them to weakly supervised learning has rarely been investigated.

\begin{figure*}[t]
	\centering
	\includegraphics[width=1\textwidth]{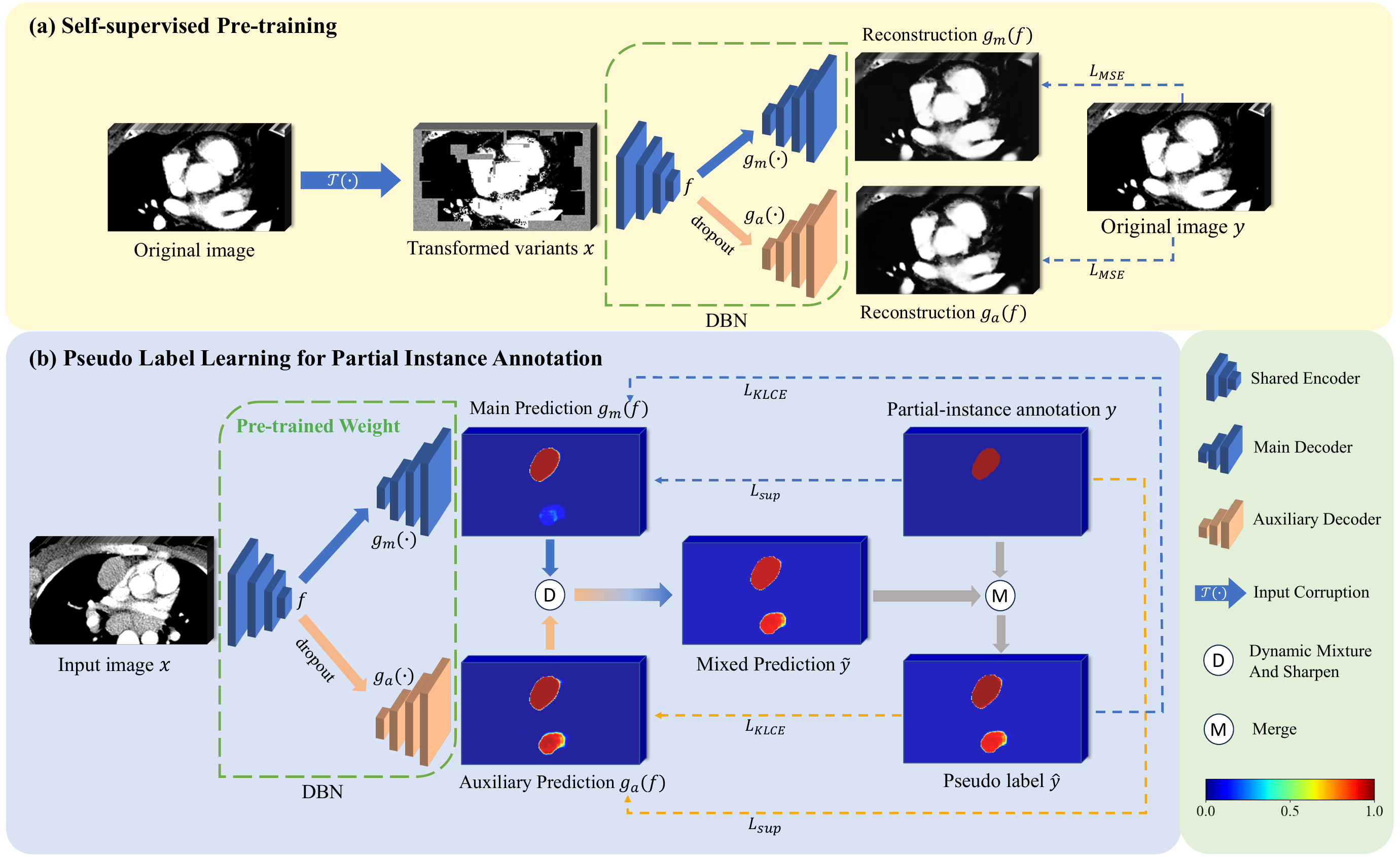}
	\caption{An overview of the proposed DBDMP which utilizes a dual-branch network with one shared encoder and two decoders. (a)~In the self-supervised learning stage, Model Genesis is employed for pre-training. (b)~In the downstream learning stage, a mixture of the outputs from the two decoders is combined with the original partial annotation to obtain a pseudo label. We also use a consensus-aware loss $\mathcal{L}_{KLCE}$ to avoid over-fitting noise in the pseudo labels.}
    \label{Pipeline}
\end{figure*}
\section{Methods}
	Fig.~\ref{Pipeline} illustrates the proposed partial instance annotation learning framework named DBDMP, which consists of a self-supervised pre-training stage and a pseudo label learning stage to deal with partial instance annotations. To achieve more stable predictions, we introduce a network with one encoder and two decoders to generate pseudo labels for unannotated instances. In self-supervised pre-training stage, as shown in Fig.~\ref{Pipeline}(a), a model with dual branches is trained separately to improve feature extraction capabilities by reconstructing corrupted images. In Fig.~\ref{Pipeline}(b), the outputs of the two decoders are mixed to obtain the pseudo label, aiming to leverage the prediction from the auxiliary branch to complement that from the main branch and supplement weak annotation information. To robustly learn the pseudo label, we introduce a consensus-aware loss function that assigns higher weights to voxels with more reliable pseudo labels. Additionally, to deal with the extreme imbalance between foreground and background voxels, we prioritize learning the foreground voxels with confidence and give lower weight to the learning of background voxels, with the aim of mining more potential lymph nodes that are unannotated.
    
    \subsection{Dual-branch Network}\label{sec:Network}
        As shown in Fig.~\ref{Pipeline}, the dual-branch network, which extends from the VNet architecture~\citep{milletari2016v}, comprises a shared encoder and two decoders inspired by~\cite{luo2022scribble}. Let's denote the inputs of the encoder as $x$ and the output features as $f$, which include features from the bottleneck layer and skip connections at different resolutions. The two decoders share the same structure, but have different inputs and parameters. The main decoder directly utilizes the features $f$ from the encoder as input, while the auxiliary decoder takes perturbed $f$ through dropout as input. We define the mappings of features from $f$ to the outputs of the main and auxiliary decoders as $g_m(\cdot)$ and $g_a(\cdot)$, respectively. In detail, one convolution block contains two convolution layers and a residual connection with Instance Normalization (IN) and Leaky ReLU activation function. Both the encoder and decoder are symmetrical structures with five different resolutions.
        
    \begin{figure*}[t]
	  \centering
	  \includegraphics[width=1\textwidth]{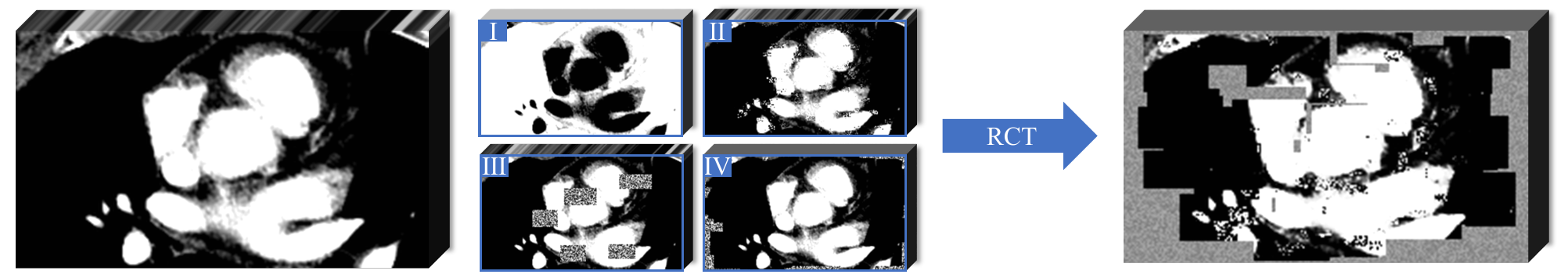}
        \caption{The transformations used to generate the input for self-supervised training: I. Non-linear transformation, II. Local pixel shuffling, III. In-painting, IV. Out-painting. RCT (Randomly Composed Transformation) means that the basic transformations are composed, each with a probability to be used. Note that in-painting and out-painting are not performed together each time.}
        \label{Model Genesis}
    \end{figure*}
    
	\subsection{Self-supervised Pre-training}\label{sec:Self training}
        An appropriate pretext task can empower neural networks to learn low-level and high-level features that are conducive to downstream tasks~\citep{jing2020self}. \cite{zhou2021models} introduced a self-supervised method called Model Genesis, which performs an image reconstruction process, and has shown promising results for downstream supervised segmentation tasks. Therefore, we use Model Genesis to pre-train the dual-branch network. Unlike the original Model Genesis that only trains one decoder, we extend it by training two decoders for the reconstruction process during pre-training.

        Model Genesis~\citep{zhou2021models} employed three types of transformations on the original images, as detailed in Fig.~\ref{Model Genesis}: 1) Non-linear transformation integrates the Bézier Curve~\citep{mortenson1999mathematics} to assign a unique determined value to each pixel, encouraging the model to focus on the information of image appearance and intensity distribution. 2) Local pixel shuffling samples a window smaller than the model's receptive field in the patch and rearranges the internal pixels to encourage the model to learn the local texture and boundary. 3) Out-painting or In-painting: Out-painting sets the outer pixels of a shape to random values, while the inner pixels retain their original intensities. In-painting follows the opposite way. The network learns visual features of images by reconstructing the original images from the corrupted version. In the self-supervised pre-training as shown in Fig.~\ref{Pipeline}(a), the outputs from the main and auxiliary decoders compute the Mean Squared Error (MSE) $\mathcal{L}_{MSE}$ separately with the original image $y$.
        The reconstruction loss is defined as: 
        \begin{align}
            \mathcal{L}_{rec} = \mathcal{L}_{MSE}(p^m, y) + \mathcal{L}_{MSE}(p^a, y)
            \label{eq:MSE}
        \end{align}
        where $p^m=g_m(f)$ and $p^a=g_a(f)$ are the predictions of the main and auxiliary decoder, respectively. And $y$ denotes the reconstruction target, i.e., the original input image.
    
    \subsection{Supervised Loss for Partial Instance Annotations}
        Partial instance annotations can be considered as a noisy label learning problem, where some foreground regions are incorrectly labeled as the background. However, compared to conventional noisy label learning scenarios, the noise in the background of partial instance annotations can be excessive, with a large amount of false negatives. Therefore, we integrate weakly supervised learning and noisy label learning methods to effectively learn from partial instance annotations and generate reliable supervision signals. This combination enables us to mitigate the impact of excessive noise in the background and produce more effective supervision signals for the learning process.

        Firstly, in our quest for more reliable supervision signals, we employ a noisy label learning technique to learn from partial instance annotation. Given a large amount of false negatives in the labels, we utilize the Symmetric Cross-Entropy (SCE) loss to balance the confidence between the partial instance annotation and the model's prediction, as proposed by~\cite{wang2019symmetric}:
        \begin{align}
            \mathcal{L}_{SCE}(p,y) = \gamma \times \mathcal{L}_{CE}(p,y) + \mathcal{L}_{CE}(y, p)
            \label{eq:SCE}
        \end{align}
        where $\mathcal{L}_{CE}$ is the widely used Cross-Entropy loss. $p$ and $y$ are the predicted probability map in a certain branch and partial instance annotation. The relationship between $\gamma$ and 1 indicates which one is more trustworthy between the prediction and the label. In this work we set $\gamma$ to 0.8 due to that the partial instance annotation is less credible than the model's prediction.
        
        Secondly, we introduce the Partial Cross-Entropy (PCE) loss~\citep{lee2020scribble2label} to ensure that the foreground in the partial instance label can be reliably learned. Unlike the PCE loss used in scribble-level annotations~\citep{luo2022scribble}, which supervises all the labeled voxels (including all target categories as well as the background) and does not calculate on unlabeled voxels, for partial instance annotations, we compute the cross-entropy only for the foreground voxels:
        \begin{align}
            \mathcal{L}_{PCE}(p,y) = - \sum_{i \in \Omega} y_i\log p_i
        \end{align}
        where $\Omega$ is the foreground voxels in partial instance annotation. $p_i$ and $y_i$ denote the predicted probability and partial instance annotation of voxel $i$.
        
        Finally, despite that $\mathcal{L}_{PCE}$ helps to improve recall of lymph nodes, it increases the risk of false positives. To deal with this problem, we additionally introduce a Tversky loss~\citep{salehi2017tversky} for supervision. Unlike Dice loss, which treats False Positives (FPs) and False Negatives (FNs) samples equally, the Tversky loss can balance the importance of both with different weights and mitigate class imbalance simultaneously:
        \begin{align}
            \mathcal{L}_{Tversky}(p,y) = \frac{TP}{TP+\alpha \times FP+(1-\alpha) \times FN}
            \label{eq:Tversky}
        \end{align}
        where $N$ is the number of voxels, $TP=\sum_{i=1}^N p_iy_i$, $FP=\sum_{i=1}^N p_i(1-y_i)$ and $FN=\sum_{i=1}^N (1-p_i)y_i$. By adjusting the hyper-parameter $\alpha$, we can control the importance between FPs and FNs. To predict more foreground voxels (false positive samples relative to partial instance annotation), $\alpha$ is set to 0.4 based on experiments.
        
        For partial instance annotation, the supervised loss for each decoder is a combination of $\mathcal{L}_{PCE}$, $\mathcal{L}_{SCE}$ and $\mathcal{L}_{Tversky}$:
        \begin{align}
            \mathcal{L}_{sup}(p,y) = \mathcal{L}_{SCE}(p, y) + \mathcal{L}_{PCE}(p, y) + \mathcal{L}_{Tversky}(p, y)
        \end{align}

    \subsection{Online Pseudo Label Learning}
        Due to the presence of incorrectly labeled background voxels in partial instance annotations, it is unreliable to directly extract supervisory signals from them. Inspired by pseudo label learning for scribble annotations~\citep{luo2022scribble}, we first dynamically mix the predictions from the two decoders: 
        \begin{align}
            \Tilde{y}_{mix} = \theta \times p^m + (1.0-\theta) \times p^a
        \end{align}
        where $\theta$ is randomly generated from a uniform distribution between 0 and 1 at each iteration, enhancing the diversity of the pseudo label and compelling the model to continually update its predictions~\citep{huo2021atso}.

        Then, we apply a sharpening function to adjust the entropy of the label distribution. The predicted probability of class $k$ can be defined as:
        \begin{align}
            \Tilde{y}^k = \frac{e^{\Tilde{y}_{mix}^k/\tau}}{\sum_{j \in C} e^{\Tilde{y}_{mix}^j/\tau}}
            \label{eq:pseudo}
        \end{align}
        where $\Tilde{y}_{mix}^k$ is the mixed output of class $k$ and $C$ is the set of all categories. $\tau$ is a temperature that is normally set to 1~\citep{hinton2015distilling}. When $\tau>1$, the labels become smoother, leading to increased entropy within the labels. Consequently, the information carried by negative labels is relatively amplified, directing the model training to pay more attention to negative labels. Conversely, when $\tau<1$, the labels become sharper. Properly sharpening the labels can enhance their robustness to noise while also maintaining the differences between classes. We set $\tau$ to 0.3 in our implementation.
        
        Finally, we integrate the mixed pseudo label with the partial instance annotation to obtain the final pseudo label, ensuring that the pseudo label complements the partial annotation. The final pseudo label is denoted as $\hat{y}$, and its $i^{th}$ element is defined as $\hat{y}_i = y_i + (1.0 - y_i)  \tilde{y}_i$, i.e., the zero region in $y$ is replaced by the corresponding values from $\tilde{y}$.

        
        
        The perturbation introduced in the auxiliary decoder may lead to uncontrollable effects. Ideally, predictions for background voxels near the classification boundary should shift towards the foreground space. However, foreground voxels in partial instance annotations may be predicted as background, leading to misleading effects in the model's training. To mitigate such adverse effects, we only learn from pixels with minor discrepancies based on the consistency of the two outputs, ensuring a smooth and gradual learning process. We utilize Kullback-Leibler (KL) divergence to estimate the consistency of the two outputs and use it to generate voxel-wise weights for the Cross-Entropy loss. Following the approach outlined in \cite{zheng2021rectifying}, the weight for voxel $i$ is defined as:
        \begin{align}
            \mathcal{W}_i=e^{-KL(p^m_i,p^a_i)}
            \label{eq:consistency_weight}
        \end{align}
        where $KL(p^m_i,p^a_i)=p^a_ilog(p^a_i/p^m_i)$ is the KL divergence loss calculated from the $i^{th}$ voxel of the two probability maps $p^m$ and $p^a$. When the predictions of a certain voxel from the main decoder and the auxiliary decoder are highly dissimilar, Eq.~\eqref{eq:consistency_weight} will lead to  a lower value of $\mathcal{W}_i$. Based on this observation, the learning loss of the main branch for pseudo label can be formulated as:
        \begin{align}
            \mathcal{L}_{p}(p^m, p^a, \hat{y}) = \frac{1}{W} \sum_{i} [\mathcal{W}_i (-\hat{y}_i\log p^m_i) + KL(p^m_i,p^a_i)]
            \label{eq:KLCE}
        \end{align}
        where $W$ is the sum of $\mathcal{W}_i$ for all voxels. The introduction of $KL$ can avoid excessive discrepancies between the predictions of the two decoders.

        The proposed DBDMP framework learns from both partial instance annotation and pseudo label by minimizing the following combined objective function:
        \begin{align}
            \mathcal{L}_{total} =& \mathcal{L}_{sup}(p^m, y) + \mathcal{L}_{sup}(p^a, y)
            \nonumber \\
            &+ \lambda_p [\mathcal{L}_{p}(p^m, p^a, \hat{y}) + \mathcal{L}_{p}(p^a, p^m, \hat{y})]
            \label{eq:total_loss}
        \end{align}
        where $y$ and $\hat{y}$ are partial instance annotation and the generated pseudo label, respectively. $\lambda_p$ is the trade-off weight that schedules with an epoch-dependent sigmoid-like ramp-up function in the first 100 epochs as the pseudo labels in the early training stage can be in poor equality:
        \begin{equation}
            \lambda_p = \lambda \times e^{-5\times(1-t/t_{max})^2}
            \label{eq:rampup}
        \end{equation}
        where $\lambda$ is a hyper-parameter that represents the final value of $\lambda_p$. $t_{max}$ is set to 99 which means the maximal epoch for ramp-up and $t$ is the current epoch.

\section{Experiments}
	\subsection{Dataset}
        \paragraph{LNQ2023 Challenge Dataset} The Mediastinal Lymph Node Quantification (LNQ): Segmentation of Heterogeneous CT Data Challenge dataset includes 513 CT volumes. Each volume contains 48 to 656 slices with slice thickness ranging from 2.0 to 5.5 $mm$ and pixel size 1.0~$mm$ $\times$ 1.0~$mm$. The matrix size in the axial plane is 512$\times$512. The images were split at patient level into 393, 20, and 100 for training, validation, and testing, respectively. In the training set, cases are partially annotated, meaning only one or several positive lymph nodes in the volumes are labeled, while all diseased lymph nodes in the validation and test sets are fully annotated.

    \begin{figure*}[t]
        \centering
        \includegraphics[width=1\textwidth]{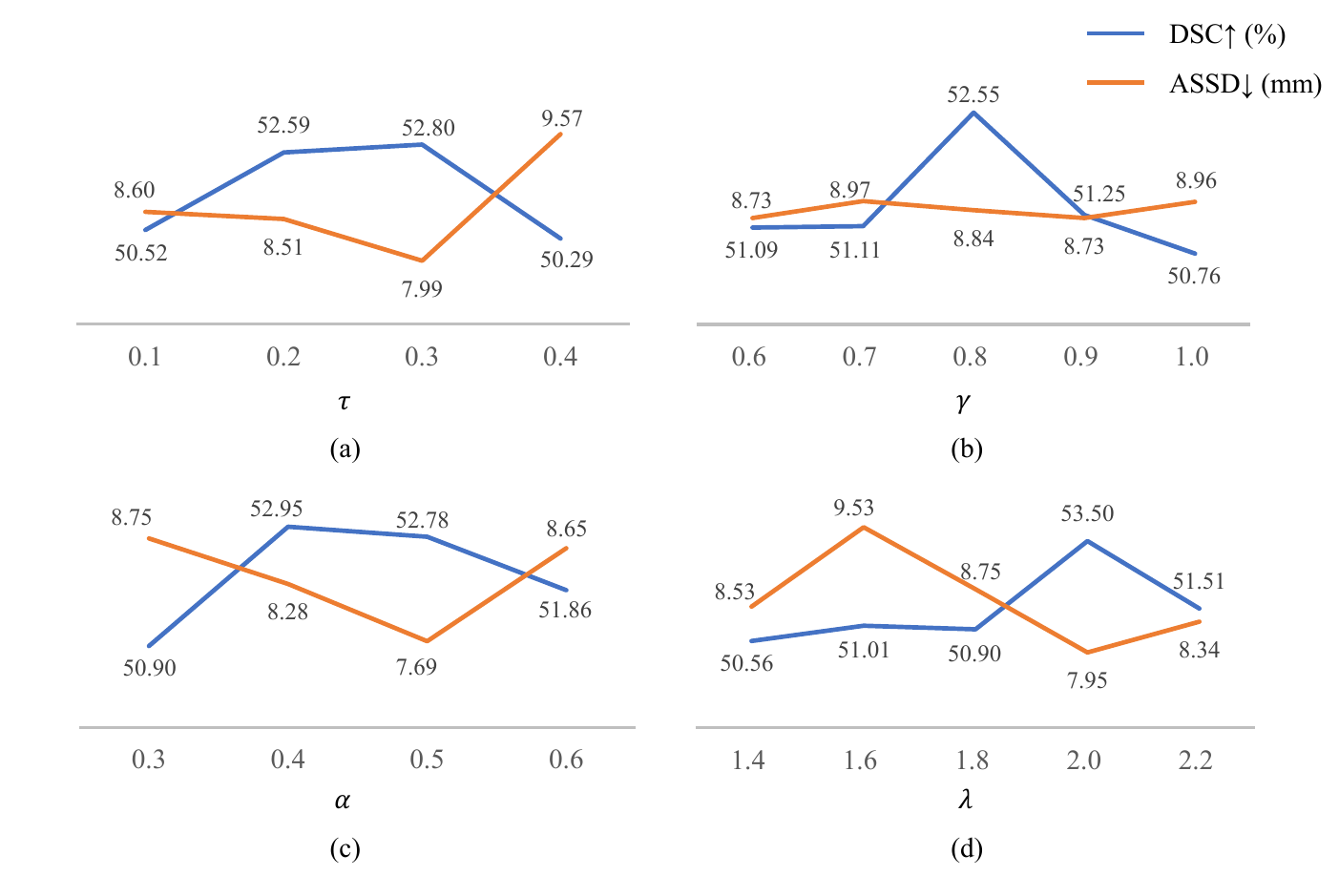}
        \caption{Sensitivity analysis of hyper-parameters $\tau$, $\gamma$, $\alpha$ and $\lambda$, respectively.}
        \label{Para Tuning}
    \end{figure*}
        
	\subsection{Implementation Details}
        Our method was implemented in nnUNet~\citep{isensee2021nnu}, which is a Pytorch-based \citep{paszke2019pytorch} toolkit for image computing with deep learning. The implementation was carried out on a single NVIDIA 2080Ti GPU with 11GB VRAM. We utilized a VNet-like~\citep{milletari2016v} network as the backbone for all experiments, and extended it to two decoders, as detailed in Section~\ref{sec:Network}.
        
        For preprocessing, we first cropped the volumes to the lung region based on intensity. Subsequently, we resampled each volume into the resolution of 3.0 $mm\times$0.8 $mm\times$0.8 $mm$. Finally, we normalized each volume to have zero mean and unit variance. Our networks were trained using a patch-based approach with a patch size of $224\times128\times64$ and a batch size of 2. We employed the polynomial learning rate strategy to decay the learning rate in each epoch.
        
        For self-supervised pre-training in Fig.~\ref{Pipeline}(a), we used Stochastic Gradient Descent (SGD) optimizer with a momentum of 0.99, an initial learning rate of 0.01, and weight decay of $3 \times 10^{-5}$ to minimize the reconstruction loss in Eq.~\eqref{eq:MSE}. The training process lasted for 1000 epochs, with 250 iterations in each epoch.
        During weakly supervised training, the segmentation model was initialized with the weights obtained from self-supervised pre-training. We minimized the loss functions in Eq.~\eqref{eq:total_loss} using SGD optimizer with a momentum of 0.9, while keeping the other parameters the same as those in the self-supervised pre-training stage. The training epoch was 300 with 250 iterations in each epoch. 
                
        During the inference stage, we loaded the weights from the final epoch and only utilized predictions from the main decoder as the final outputs. All inference processes were conducted using a sliding window strategy. We also applied a specific post-processing method, which involved removing lymph node regions at the boundaries of an image and eliminating a portion of the lymph nodes based on voxel intensity and the actual volume. For quantitative evaluation, we calculated the Dice Similarity Coefficient (DSC) and the Average Symmetric Surface Distance (ASSD) between a segmentation result and the ground truth. In light of the potential for samples with failed predictions, the ASSD for these samples is missing, so we fill them with the maximum value of the successfully calculated ASSD and then average them. This may result in larger mean ASSD values.

    \begin{table*}[t]
        \centering
        \caption{Ablation study of our proposed method on LNQ2023 Challenge Dataset for both validation set and test set. DBN is the Dual-branch Network detailed in Section~\ref{sec:Network}. DBN$\dagger$ means using the pre-training weight generated by Model Genesis~\citep{zhou2021models} in Section~\ref{sec:Self training}. $\mathcal{L}_{Dice}$ is the widely used Dice loss and $\mathcal{L}_{KLCE}$ is explained in detail in Eq.~\eqref{eq:KLCE}. Training VNet with one encoder and one decoder by $\mathcal{L}_{CE}$ is served as the baseline method.}
        \resizebox{1\textwidth}{!}{
            \begin{tabular}{cccc|cc|cc}
            \hline
            \multirow{2}[4]{*}{} & \multicolumn{1}{c}{\multirow{2}[4]{*}{Network}} & \multirow{2}[4]{*}{$\mathcal{L}_{sup}$} & \multirow{2}[4]{*}{$\mathcal{L}_{p}$} & \multicolumn{2}{c|}{Validation Set} & \multicolumn{2}{c}{Test Set} \bigstrut[t]\\
            \cline{5-8}      &       &       &       & \multicolumn{1}{l}{DSC$\uparrow$(\%)} & \multicolumn{1}{l|}{ASSD$\downarrow$(mm)} & \multicolumn{1}{l}{DSC$\uparrow$(\%)} & \multicolumn{1}{l}{ASSD$\downarrow$(mm)} \bigstrut\\
            \hline
            (a) & DBN & $\mathcal{L}_{CE}$ & $\mathcal{L}_{Dice}$ & 15.12$_{\pm{14.75}}$ & 26.94$_{\pm{11.78}}$ & 23.22$_{\pm{17.50}}$ & 25.41$_{\pm{15.33}}$ \bigstrut\\
            (b) & DBN & $\mathcal{L}_{CE}+\mathcal{L}_{Tversky}$ & $\mathcal{L}_{Dice}$ & 34.99$_{\pm{25.82}}$ & 16.39$_{\pm{12.73}}$ & 45.10$_{\pm{21.58}}$ & 13.54$_{\pm{12.05}}$ \\
            (c)  & DBN & $\mathcal{L}_{CE}+\mathcal{L}_{Tversky}$ & $\mathcal{L}_{KLCE}$ & 32.10$_{\pm{24.29}}$ & 15.88$_{\pm{9.09}}$ & 40.26$_{\pm{22.39}}$ & 15.19$_{\pm{10.94}}$ \\
            (d)  & DBN & $\mathcal{L}_{SCE}+\mathcal{L}_{Tversky}$ & $\mathcal{L}_{KLCE}$  & 31.48$_{\pm{25.68}}$ & 17.63$_{\pm{11.84}}$ & 39.34$_{\pm{22.58}}$ & 16.54$_{\pm{15.75}}$ \\
            (e)  & DBN & $\mathcal{L}_{PCE}+\mathcal{L}_{Tversky}$ & $\mathcal{L}_{KLCE}$  & 52.53$_{\pm{22.29}}$ & \underline{8.39}$_{\pm{6.84}}$ & \textbf{57.36}$_{\pm{17.09}}$ & \underline{9.85}$_{\pm{13.25}}$ \\
            (f)  & DBN & $\mathcal{L}_{PCE}+\mathcal{L}_{SCE}+\mathcal{L}_{Tversky}$ & $\mathcal{L}_{KLCE}$  & \underline{53.31}$_{\pm{20.40}}$ & \textbf{8.07}$_{\pm{6.55}}$ & \underline{56.10}$_{\pm{17.58}}$ & 10.28$_{\pm{12.75}}$ \\
            (g)  & DBN$\dagger$ & $\mathcal{L}_{PCE}+\mathcal{L}_{SCE}+\mathcal{L}_{Tversky}$ & $\mathcal{L}_{KLCE}$  & \textbf{54.10}$_{\pm{21.92}}$ & 8.72$_{\pm{7.71}}$ & 55.44$_{\pm{18.98}}$ & \textbf{9.35}$_{\pm{7.69}}$ \bigstrut[b]\\
            \hline
            Baseline & VNet & \multicolumn{2}{c|}{$\mathcal{L}_{CE}$} & 11.04$_{\pm{17.86}}$ & 20.83$_{\pm{8.64}}$ & 19.04$_{\pm{19.40}}$ & 18.23$_{\pm{10.10}}$ \bigstrut\\
            \hline
            \end{tabular}%
            }
        \label{tab:Ablation Study}%
    \end{table*}%
    
    \begin{figure*}[t]
        \centering
        \includegraphics[width=\textwidth]{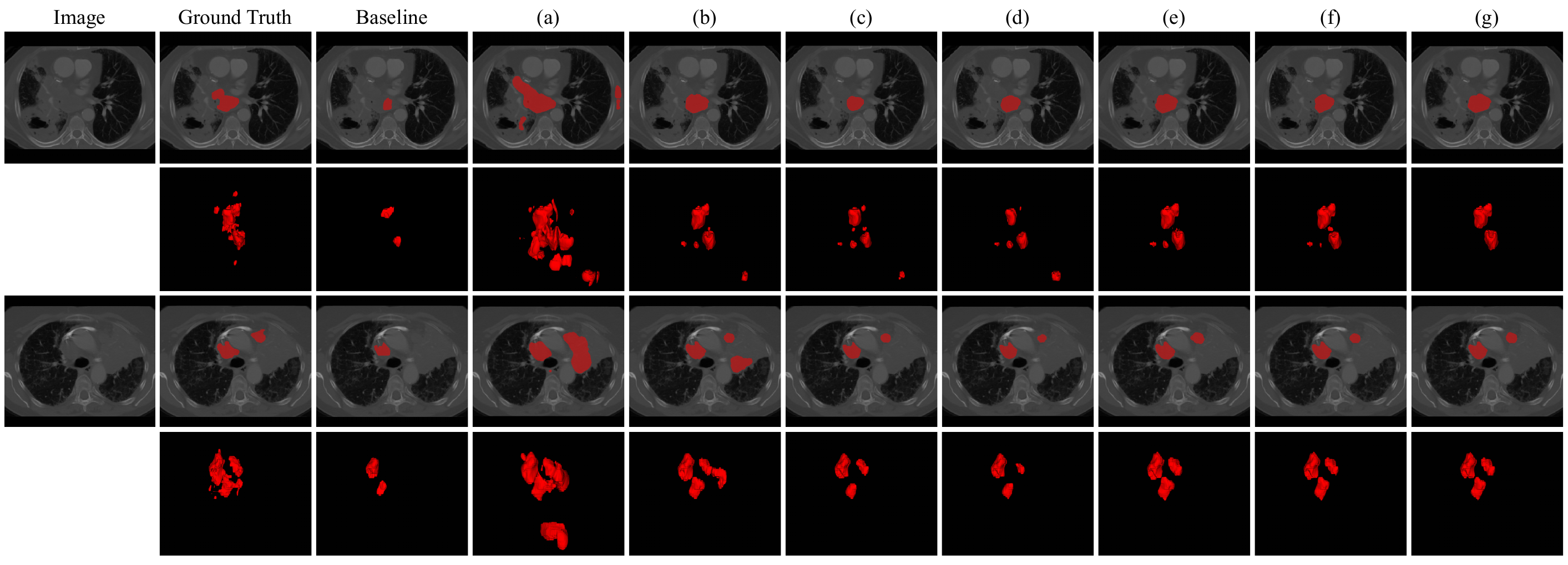}
        \caption{Visualization of segmentation results of the ablation study in Table \ref{tab:Ablation Study}.}
        \label{visualization}
    \end{figure*}
    
    \subsection{Results}
        \paragraph{Sensitivity Analysis of Some Hyper-parameters} We conducted experiments to evaluate the sensitivity of the hyper-parameters $\tau$ in Eq.~\eqref{eq:pseudo}, $\gamma$ in Eq.~\eqref{eq:SCE}, $\alpha$ in Eq.~\eqref{eq:Tversky} and $\lambda$ in Eq.~\eqref{eq:rampup}. Fig.~\ref{Para Tuning} presents the results obtained on the validation set.
        
        The hyper-parameter $\tau$ governs the extent of sharpening applied to the soft pseudo labels. We investigated the segmentation performance of the proposed framework by setting $\tau$ to 0.1, 0.2, 0.3, and 0.4, respectively. As illustrated in Fig.~\ref{Para Tuning}(a), when $\tau$ increases from 0.1 to 0.3, the DSC score improves. However, the performance is decreased when $\tau$ increases to 0.4, showing that the best value of $\tau$ is 0.3.
        
        The hyper-parameter $\gamma$ signifies the degree of confidence between the model's predictions and partial instance annotations. The results depicted in Fig.~\ref{Para Tuning}(b) indicate that, considering both DSC and ASSD, the model achieved the best result when $\gamma=$~0.8. This suggests that the model deems its own predictions more reliable than the partial instance annotations.
        
        The hyper-parameter $\alpha$ in Eq.~\eqref{eq:Tversky} balances the penalty imposed on False Positives (FPs) and False Negatives (FNs), and was tested with different values in \{0.3, 0.4, 0.5, 0.6\}. When $\alpha$ is smaller, the model imposes less penalty on FPs, thus encouraging the model to predict more positive results than those in the partial instance annotations. However, a weaker penalty on FPs may lead to an over-prediction of foreground voxels. In Fig.~\ref{Para Tuning}(c), $\alpha=$~0.4 and $\alpha=$~0.5 achieved very close DSC scores, and $\alpha=$~0.5 has a lower ASSD values. When  $\alpha=$~0.3, the performance was much lower. The increased ASSD may be due to that the model predicted some foreground voxels that are far from the actual lymph nodes.
        
        The value of $\lambda$ represents the confidence in the quality of the generated pseudo labels during the training process. We conducted experimental tests with the set \{1.4, 1.6, 1.8, 2.0, 2.2\}. As illustrated in Fig.~\ref{Para Tuning}(d), the result is notably superior when $\lambda=$~2.0 compared to other settings.
        
    \paragraph{Ablation Study} We conducted additional experiments to validate the effectiveness of Dual-Branch Network (DBN) and the modifications made to adapt the work of \cite{luo2022scribble} for learning from partial instance annotations. The quantitative results on the validation set and test set are presented in Table.~\ref{tab:Ablation Study}, where the baseline method was taking the partial annotations as full ones to train a VNet with cross entropy loss. 

    Table~\ref{tab:Ablation Study} shows that the baseline method only achieved an average Dice of 11.04\% on the validation set, indicating  insufficient supervision from the partial annotations. By leveraging pseudo labels from the dual-branch network with $\mathcal{L}_{Dice}$, it was improved to 15.12\%. By additionally introducing the Tversky loss for the original partial annotations, the average Dice was 34.99\%.  By combing $\mathcal{L}_{PCE} + \mathcal{L}_{SCE} + \mathcal{L}_{Tversky}$ for partial instance annotations and $L_{KLCE}$ for pseudo labels, the average Dice was 53.31\%, and leveraging a pretrained model based on Model Genesis further improved it to 54.10\%, showing the effectiveness the loss design and pretraining strategy of our method.  
    
    On the testing set, comparison between (c) and (e) in Table~\ref{tab:Ablation Study} shows that replacing $\mathcal{L}_{CE}$ by $\mathcal{L}_{PCE}$ substantially improved the average Dice from 40.26\% to 57.36\%, showing the effectiveness of reducing the contribution of background voxels in the loss calculation when many lymph nodes are incorrectly labeled as the background in parial instance annotations. 
    The proposed method achieved the lowest ASSD value of 9.35~$mm$. Despite that using $\mathcal{L}_{PCE} + \mathcal{L}_{SCE} + \mathcal{L}_{Tversky}$ for partial instance annotations achieved a lower DSC value than  $\mathcal{L}_{PCE} + \mathcal{L}_{Tversky}$, their gap is relatively small considering the performance of the other methods. The different performance between validation set and  testing set is mainly from the data distribution shift between the two subsets.   

    Fig.~\ref{visualization} shows a visual comparison between the compared methods listed in Table~\ref{tab:Ablation Study}. It can be observed that the baseline method has obvious under-segmentation, due to taking false negatives in the partial annotation as the background. The naive pseudo label learning method (a) has a lot of over-segmentation, due to that the pseudo labels contain many false positives. By using $L_{KLCE}$ for the pseudo labels and $L_{PCE}$, $L_{SCE}$, $L_{Tversky}$ for partial annotations respectively, the performance continues to improve.
        

\section{Discussion and Conclusion}
	In this study, we explored a weakly supervised learning framework based on partial instance annotations for lymph node segmentation. 
    For such annotations, the key is to identify trustworthy background regions and provide strong foreground signals during training to ensure robust learning of foreground voxels. Our method deals with this problem by generating pseudo labels to mine more potential lymph nodes. As pseudo labels from a single prediction branch may have bias, we propose dynamic mixture of predictions from two branches, leading to more stable pseudo labels and better uncertainty estimation of them based on divergence between the two branches. Loss functions are also carefully designed to highlight the foreground class  while reducing the effect of noise in pseudo labels. 

    This work also has some limitations that could be addressed in the future. First, the segmentation model in our work only learns from the LNQ dataset, and the performance may be further improved by leveraging other existing fully supervised datasets. We believe our approach is adaptable enough for mixed datasets that have both partially and fully annotated cases. Second, the LNQ dataset has only large lymph nodes labeled in the training set, and the distributions of the labeled ones for training and those for testing may be different, making it more challenging to obtain robust performance during testing, especially for small lymph nodes that have not been annotated in the training set. Improving the diversity of the labeled cases under the same annotation budget is a potential solution for this problem, such as making the labeled cases contain lymph nodes with different scales, positions, and shapes. In addition, the loss function in  this work has several hyper-parameters, and they are searched manually. In the future, it would be interesting to automate the determination of these hyper-parameters.
 
    

    In conclusion, we proposed a partial instance annotation learning framework that combines  weakly supervised learning and noisy label learning for lymph node segmentation. By introducing a dual-branch network, we dynamically mixed the outputs from the two decoders and fused them with partial instance annotations to obtain reliable pseudo labels. In learning from partial instance annotations, the introduction of multiple loss functions not only provides more reliable foreground and background supervision signals but also facilitates the segmentation of potential lymph nodes that are not labeled out. We conducted experiments using the dataset from the Mediastinal Lymph Node Quantification Challenge, without using any other datasets for pre-training or during the training stage. We finally achieved an average DSC of 54.10\% and 55.44\%, and average ASSD of 8.72~$mm$ and 9.35~$mm$ on validation set and test set, respectively. In the future, it is of interest to leverage other labeled or unlabeled datasets to assist the learning process, such as using unannotated datasets for self-supervised pre-training, or leveraging a small number of fully labeled images to boost the segmentation performance. 


\acks{This work was supported by National Natural Science Foundation of China under grant 62271115, Radiation Oncology Key Laboratory of Sichuan Province Open Found under grant 2022ROKF04, and Science and Technology Department of Sichuan Province under grant 2022YFSY0055.}

%
\ethics{The work follows appropriate ethical standards in conducting research and writing the manuscript, following all applicable laws and regulations regarding treatment of animals or human subjects.}

\coi{We declare we do not have conflicts of interest.}

\data{All data used in the paper are from the Mediastinal Lymph Node Quantification (LNQ): Segmentation of Heterogeneous CT Data competition. The training set data and annotations, as well as the validation set data, can be obtained from the competition website \url{https://lnq2023.grand-challenge.org/}. However, annotations for the validation set and both the data and annotations for the test set are not available on the website.}

\bibliography{sample}





\end{document}